\def\year{2016}\relax
\documentclass[letterpaper]{article}
\usepackage{aaai16}
\usepackage{times}
\usepackage{amsmath}
\usepackage{amssymb}
\usepackage{comment}
\usepackage{titling}

\usepackage{grfext}
\PrependGraphicsExtensions*{.pdf}
\usepackage{graphicx}
\usepackage{authblk}
\title{Hierarchical Linearly-Solvable Markov Decision Problems\\(Extended Version with Supplementary Material)}
\author{{\Large\textbf{Anders Jonsson}}}
\author{{\Large\textbf{Vicen\c{c} G\'omez}}}
\affil{Department of Information and Communication Technologies\\
Universitat Pompeu Fabra\\
Roc Boronat 138, 08018 Barcelona, Spain\\
\{anders.jonsson,vicen.gomez\}@upf.edu
}
\date{}
\begin{document}

%

\maketitle

\begin{abstract}
We present a hierarchical reinforcement learning framework that formulates each task in the hierarchy as a special type of Markov decision process for which the Bellman equation is linear and has analytical solution. Problems of this type, called linearly-solvable MDPs (LMDPs) have interesting properties that can be exploited in a hierarchical setting, such as efficient learning of the optimal value function or task compositionality. The proposed hierarchical approach can also be seen as a novel alternative to solving LMDPs with large state spaces. We derive a hierarchical version of the so-called Z-learning algorithm that learns different tasks simultaneously and show empirically that it significantly outperforms the state-of-the-art learning methods in two classical hierarchical reinforcement learning domains: the taxi domain and an autonomous guided vehicle task.
\end{abstract}

\section{Introduction} 
Hierarchical reinforcement learning (HRL) is a general framework for addressing large-scale reinforcement learning problems.
It exploits the task (or action) structure of a problem by considering policies over temporally extended actions that typically involve a reduced subset of the state components.
For example, the MAXQ approach  \cite{dietterich2000hierarchical} decomposes a Markov decision process (MDP) 
and its value function into a hierarchy of smaller MDPs such that the value function of the target MDP corresponds to an additive combination
of the value functions of the smaller MDPs.
Another example is the options approach for which different tasks can be learned simultaneously in an online fashion
\cite{Sutton98intra-optionlearning}.
HRL methods have also been used to explain human and animal behavior~\cite{botvinick2009hierarchically}.

Independently, a class of stochastic optimal control problems was introduced for which the actions and cost function
are restricted in ways that make the Bellman equation linear and thus more efficiently solvable \cite{TodorovNIPS2007,Kappen2005}.
This class of problems is known in the discrete setting as linearly-solvable MDPs (LMDPs), in the continuous setting as path-integral control or more generally, as Kullback-Leibler (KL) control~\cite{KappenML2012}.
Optimal control computation for this class of problems is equivalent to a KL-divergence minimization.

The original LMDP formulation considers a single action that changes the stochastic laws of the environment. An alternative interpretation (that we adopt in this work) is to consider a stochastic policy over deterministic actions.
LMDPs have many interesting properties. For example, optimal control laws for LMDPs can be linearly combined to derive composite optimal control laws efficiently~\cite{TodorovNIPS2009}.
Also, the power iteration method, used to solve LMDPs, is equivalent 
to the popular belief propagation algorithm used for probabilistic inference in dynamic graphical models~\cite{KappenML2012}.
The optimal value function for LMDPs can be learned efficiently using an off-policy learning algorithm, Z-learning, that operates directly in the state space instead of in the product space of states and actions \cite{TodorovPNAS2009}.

In the continuous setting, the KL-divergence reduces to the familiar quadratic energy cost,
widely used in robotic applications. Examples of such applications include robot navigation~\cite{KinjoFN2013} and motor skill reinforcement learning~\cite{Theodorou2010}. 
This class of problems is also relevant in other disciplines, such as cognitive science and decision making theory \cite{friston2013anatomy,ortega2013thermodynamics}.
However, in general, application of LMDPs to real-world problems is challenging, mainly due to the curse dimensionality~\cite{Abbasi,taka2014,TodorovADPRL2009}.


In this paper, we propose to combine both HRL and LMDP frameworks and formulate a reinforcement learning problem as a hierarchy of LMDPs.
Surprisingly, despite LMDPs were introduced already ten years ago, no unifying framework that combines both methodologies has been proposed yet.
The benefits of this combination are two-fold. On one hand, HRL problems expressed in this way can benefit from the same properties that LMDPs enjoy.
For example, one can use Z-learning as an efficient alternative to the state-of-the-art HRL methods.
Another example is task compositionality, by which a composite task can be learned at no cost given the optimal solution for the different composing tasks.
This is useful in tasks that have several terminal states, as we will show later.
On the other hand, LMDPs can also benefit from the HRL framework, for example, by addressing the curse of dimensionality in an alternative way
to the previously mentioned approaches or by simultaneous intra-task learning from HRL.

The paper is organized as follows. We review HRL and LMDPs in Section~\ref{sec:prel}.
The main contribution of this work is the hierarchical formulation for LMDPs, which we present in Section~\ref{sec:hlmpds}.
We empirically illustrate its benefits on two benchmarks in Section~\ref{sec:exp}~~We conclude this work in Section~\ref{sec:end}.
%
%

\section{Preliminaries}\label{sec:prel}

In this section we introduce preliminaries and notation. We first define MDPs and Semi-MDPs, then explain the idea behind MAXQ decomposition, and finally describe linearly-solvable MDPs.

\subsection{MDPs and Semi-MDPs}

An MDP $M=\langle S,A,P,R\rangle$ consists of a set of states $S$, a set of actions $A$, a transition probability distribution $P:S\times A\times S\rightarrow[0,1]$ satisfying $\sum_{s'}P(s'|s,a)=1$ for each state-action pair $(s,a)\in S\times A$, and an expected reward function $R:S\times A\rightarrow\mathbb{R}$. The aim is to learn an optimal policy $\pi:S\rightarrow A$, i.e.~a mapping from states to actions that maximizes expected future reward.

MDPs are usually solved by defining a value function $V:S\rightarrow\mathbb{R}$ that estimates the expected future reward in each state. In the undiscounted case, the optimal value is obtained by solving the Bellman optimality equation:
\begin{align*}
V(s)&=\max_{a\in A}\left\{R(s,a)+E_{s'}\left[V(s')\right]\right\}\\
&=\max_{a\in A}\left\{R(s,a)+\sum_{s'}P(s'|s,a)V(s')\right\}.
\end{align*}
To bound the optimal value function, we only consider {\em first exit} problems that define a set of terminal states $T\subseteq S$. A function $g:T\rightarrow\mathbb{R}$ defines the final reward $V(t)\equiv g(t)$ of each terminal state.
As an alternative to the value function $V$, one can define an action-value function $Q:S\times A\rightarrow\mathbb{R}$ that estimates the expected future reward for each state-action pair $(s,a)\in S\times A$.

The Bellman optimality equation can be solved globally using algorithms such as value iteration and policy iteration. However, for large state spaces this is not feasible. Alternative algorithms make local updates to the value function online. Arguably, the most popular online algorithm for MDPs is Q-learning \cite{Watkins:1989}. Given a transition $(s_t,a_t,r_t,s_{t+1})$ from state $s_t$ to state $s_{t+1}$ when taking action $a_t$ and receiving reward $r_t$, Q-learning makes the following update to the estimate $\hat{Q}$ of the optimal action-value function:
\begin{align*}
\hat{Q}(s_t,a_t)\leftarrow(1-\alpha)\hat{Q}(s_t,a_t)+\alpha(r_t+\max_a\hat{Q}(s_{t+1},a)),
\end{align*}
where $\alpha$ is a learning rate.

A Semi-MDP generalizes an MDP by including actions that take more than one time-step to complete. In this case, the Bellman optimality equation becomes
\begin{align*}
V(s)&=\max_{a\in A}E_\tau\left\{R(s,a,\tau)+\sum_{s'}P(s',\tau|s,a)V(s')\right\}\\
&=\max_{a\in A}\sum_\tau\sum_{s'}P(s',\tau|s,a)\left\{R(s,a,\tau)+V(s')\right\},
\end{align*}
where $\tau$ is the duration of action $a$, $R(s,a,\tau)$ is the expected reward when $a$ applied in $s$ lasts for $\tau$ steps, and $P(s',\tau|s,a)$ is the probability of transitioning to $s'$ in $\tau$ steps. By defining $\overline{R}(s,a)\equiv\sum_{\tau,s'}P(s',\tau|s,a)R(s,a,\tau)$ and $\overline{P}(s'|s,a)=\sum_\tau P(s',\tau|s,a)$, we get
\begin{align*}
V&(s)=\max_{a\in A}\left\{\overline{R}(s,a)+\sum_{s'}\overline{P}(s'|s,a)V(s')\right\},
\end{align*}
i.e.~a Semi-MDP can be treated and solved as an MDP.

\subsection{MAXQ Decomposition}

MAXQ decomposition \cite{dietterich2000hierarchical} decomposes an MDP $M=\langle S,A,P,R\rangle$ into a finite set of tasks $\mathcal{M}=\{M_0,\ldots,M_n\}$ with root task $M_0$, i.e.~solving $M_0$ is equivalent to solving $M$. Each task $M_i=\langle T_i,A_i,\tilde{R}_i\rangle$, $0\leq i\leq n$, consists of a termination set $T_i\subset S$, an action set $A_i\subset\mathcal{M}$ and a pseudo-reward $\tilde{R}_i:T_i\rightarrow\mathbb{R}$. 

A task $M_i$ is primitive if it has no subtasks, i.e.~if $A_i=\emptyset$. A primitive task $M_i$ corresponds to an action $a\in A$ of the original MDP $M$, defined such that $M_i$ is always applicable, terminates after one time step and has pseudo-reward $0$ everywhere. A non-primitive task $M_i$ can only be applied in non-terminal states (i.e.~states not in $T_i$). Terminating in state $t\in T_i$ produces pseudo-reward $\tilde{R}_i(t)$. $M_i$ corresponds to a Semi-MDP with action set $A_i$, i.e.~actions are other tasks.

\begin{figure}[t]
  \begin{center}
  \includegraphics[scale=1]{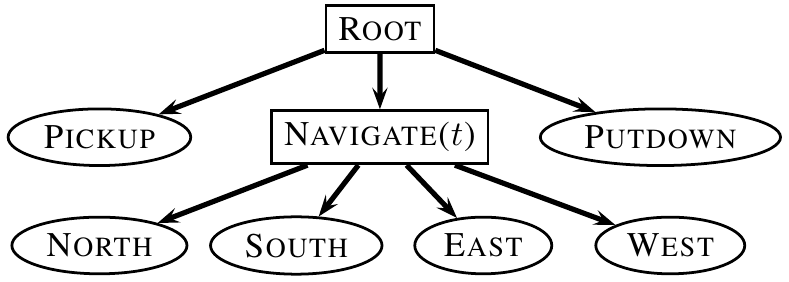}
  \end{center}

  \caption{The task graph of the Taxi domain.}
  \label{fig:taxi}
\end{figure}

MAXQ defines a {\em task graph} with tasks in $\mathcal{M}$ as nodes. There is an edge between nodes $M_i$ and $M_j$ if and only if $M_j\in A_i$, i.e.~if $M_j$ is an action of task $M_i$. To avoid infinite recursion, the task graph has to be acyclic.
Figure~\ref{fig:taxi} shows a simplified task graph of the Taxi domain, commonly used to illustrate MAXQ decomposition.

The aim of MAXQ decomposition is to learn a hierarchical policy $\pi=(\pi_0,\ldots,\pi_n)$, i.e.~a separate policy $\pi_i:S\rightarrow A_i$ for each individual task $M_i$, $0\leq i\leq n$. Each task $M_i$ defines its own value function $V_i$ that, for each state $s\in S$, estimates the expected cumulative reward until $M_i$ terminates. The reward associated with applying action $M_j\in A_i$ in state $s$ of task $M_i$ equals the value of $M_j$ in $s$, i.e.~$R_i(s,M_j)=V_j(s)$. Hence the Bellman optimality equation for $M_i$ decomposes as
\begin{align*}
V_i(s)&=\max_{M_j\in A_i}\left\{V_j(s)+\sum_{s'}P(s'|s,M_j)V_i(s')\right\},
\end{align*}
where $P(s'|s,M_j)$ is the probability of transitioning from $s$ to $s'$ when applying the (possibly composite) action $M_j$. If $M_j$ is a primitive task corresponding to an action $a\in A$ of the original MDP $M$, its value is the expected immediate reward, i.e.~$V_j(s)=R(s,a)$. The pseudo-reward $\tilde{R}_i$ is only used for learning the policy $\pi_i$ of $M_i$ and does not contribute to the value function.


\citeauthor{dietterich2000hierarchical} (\citeyear{dietterich2000hierarchical}) proposed an online algorithm for MAXQ decomposition called MAXQ-Q learning. The algorithm maintains two value functions for each task $M_i$: an estimate $\hat{V}_i$ of the value function $V_i$ defined above, and an estimate $\tilde{V}_i$ of the expected cumulative reward that includes the pseudo-reward $\tilde{R}_i$. The estimate $\tilde{V}_i$ defines the policy $\pi_i$ for $M_i$, while the estimate $\hat{V}_i$ is passed as reward to parent tasks of $M_i$. MAXQ-Q learning achieves {\em recursive optimality}, i.e.~each policy $\pi_i$ is locally optimal with respect to $M_i$. \citeauthor{conf/nips/Dietterich99} (\citeyear{conf/nips/Dietterich99}) also showed how to use state abstraction to simplify learning in MAXQ decomposition.

\subsection{Linearly-Solvable MDPs}
\label{sec:lmdps}
Linearly-solvable MDPs (LMDPs) were first introduced by \citeauthor{TodorovNIPS2007} (\citeyear{TodorovNIPS2007,TodorovPNAS2009}). The original formulation has no explicit actions, and control consists in changing a predefined uncontrolled probability distribution over next states. An alternative interpretation is to view the resulting probability distribution as a stochastic policy over deterministic actions. 
Todorov's idea was to transform the discrete optimization problem over actions to a continuous optimization problem over transition probabilities, which is convex and analytically tractable.

Formally, an LMDP $L~=~\langle S,P,R\rangle$ consists of a set of states $S$, an uncontrolled transition probability distribution $P:S\times S\rightarrow[0,1]$ satisfying $\sum_{s'}P(s'|s)=1$ for each state $s\in S$, and an expected reward function $R:S\rightarrow\mathbb{R}$. Given a state $s\in S$ and any next state distribution $D$, we define the set of next states under $D$ as $N(s,D)=\{s':D(s'|s)>0\}$.
For first-exit problems, LMDPs also have a subset of terminal states $T\subset S$.


The control in LMDPs is a probability distribution $a(\cdot|s)$ over next states;
for a given next state $s'\in S$, $a(s'|s)$ can be non-zero only if $P(s'|s)$ is non-zero. The reward $\mathcal{R}(s,a)$ for applying control $a$ in state $s$ is
\begin{align*}
\mathcal{R}(s,a)&=R(s)-\lambda\cdot\mathrm{KL}(a(\cdot|s)\Vert\, P(\cdot|s))\\
&=R(s)-\lambda\cdot E_{s'\sim a(\cdot|s)}\left[\log\frac{a(s'|s)}{P(s'|s)}\right],
\end{align*}
where $R(s)$ is the (non-positive) reward associated with state $s$.
$\mathrm{KL}(a(\cdot|s)\Vert\, P(\cdot|s))$ is the Kullback-Leibler divergence between $a$ and $P$, penalizing controls that are significantly different from $P$.
Typically, $P$ is a random walk and $\lambda$ acts as a temperature parameter.
Large values of $\lambda$ (high temperature) lead to solutions which are more stochastic, since deviating from the random dynamics is penalized more.
Conversely, very small values of $\lambda$ (low temperature) result in deterministic policies, since the state-dependent term dominates the immediate cost.
LMDPs, in a sense, replace deterministic policies defined over stochastic actions with stochastic policies defined over deterministic actions.
In what follows, unless otherwise stated, the next state $s'$ is always drawn from the distribution $a(\cdot|s)$.

We can now define the Bellman optimality equation:
\begin{align*}
\frac{1}{\lambda}V(s)&=\frac{1}{\lambda}\max_{a\in\mathcal{A}(s)}\left\{\mathcal{R}(s,a)+E_{s'}\left[V(s')\right]\right\}\\
&=\frac{1}{\lambda}R(s)+\max_{a\in\mathcal{A}(s)}E_{s'}\left[\frac{1}{\lambda}V(s')-\log\frac{a(s'|s)}{P(s'|s)}\right].
\end{align*}
For a given state $s\in S$, the set $\mathcal{A}(s)$ consists of control inputs that satisfy $\sum_{s'}a(s'|s)=1$ and $a(s'|s)>0\rightarrow P(s'|s)>0$ for each $s'\in S$.
To bound the values $V(s)$ in the absense of a discount factor, terminal states are absorbing, i.e.~$P(t|t)=1$ for each $t\in T$.

Introducing $Z(s)=e^{V(s)/\lambda}$ we obtain
\begin{align*}
\frac{1}{\lambda}V(s)&=\frac{1}{\lambda}R(s)+\max_aE_{s'}\left[-\log\frac{a(s'|s)}{P(s'|s)Z(s')}\right]\\
&=\frac{1}{\lambda}R(s)-\min_aE_{s'}\left[\log\frac{a(s'|s)}{P(s'|s)Z(s')}\right].
\end{align*}
To obtain a KL divergence on the right-hand side, introduce a normalization term $\mathcal{G}[Z](s)=\sum_{s'}P(s'|s)Z(s')$ and insert it into the Bellman equation:
\begin{align*}
\frac{1}{\lambda}V(s)=&\frac{1}{\lambda}R(s)-\min_aE_{s'}\left[\log\frac{a(s'|s)\mathcal{G}[Z](s)}{P(s'|s)Z(s')\mathcal{G}[Z](s)}\right]\\
=&\frac{1}{\lambda}R(s)+\log\mathcal{G}[Z](s)\\
&-\min_a\mathrm{KL}\left(a(\cdot|s)\middle\Vert\frac{P(\cdot|s)Z(\cdot)}{\mathcal{G}[Z](s)}\right).
\end{align*}
The KL term achieves a minimum of $0$ when the distributions are equal, i.e.~the optimal policy is
\[
a^*(s'|s)=\frac{P(s'|s)Z(s')}{\mathcal{G}[Z](s)}.
\]
Exponentiating the Bellman equation gives 
\begin{align*}
Z(s)=e^{R(s)/\lambda}\mathcal{G}[Z](s).
\end{align*}
We can write this equation in matrix form as
\begin{equation}\label{eq:lmdp}
{\bf z}=\Omega\Pi{\bf z},
\end{equation}
where $\Omega$ is a diagonal matrix with the terms $e^{R(s)/\lambda}$ on the diagonal and $\Pi$ is the transition probability matrix derived from the distribution $P$. Unlike the Bellman optimality equation, this is a system of linear equations.

Since Equation~\eqref{eq:lmdp} is linear, we can solve its eigenvector problem using, for example, the power iteration method. As an alternative, \citeauthor{TodorovNIPS2007} (\citeyear{TodorovNIPS2007,TodorovPNAS2009}) proposed an online learning algorithm for LMDPs called Z-learning. Similar to Q-learning for MDPs, the idea of Z-learning is to follow a trajectory, record transitions and perform incremental updates to the value function.

Since LMDPs have no explicit actions, each transition $(s_t,r_t,s_{t+1})$ consists of a state $s_t$, a next state $s_{t+1}$ and a reward $r_t$ recorded during the transition. Z-learning maintains an estimate $\hat{Z}$ of the optimal $Z$ value, and this estimate is updated after each transition as
\begin{align}\label{eqn:zlearning}
\hat{Z}(s_t) \leftarrow (1-\alpha)\hat{Z}(s_t) + \alpha e^{r_t/\lambda}\hat{Z}(s_{t+1}),
\end{align}
where $\alpha$ is a learning rate.


Naive Z-learning samples transitions from the passive dynamics $P$, which essentially amounts to a random walk and leads to slow learning. A better alternative is to use {\em importance sampling} to guide exploration by sampling transitions from a more informed distribution. A natural choice is the estimated optimal policy $\hat{a}$ derived from $\hat{Z}$, resulting in the following corrected update rule~\cite{TodorovNIPS2007}:
\begin{align}\label{eqn:zlearning-imp}
\hat{Z}(s_t) \leftarrow (1-\alpha)& \hat{Z}(s_t) + \alpha e^{r_t/\lambda}\hat{Z}(s_{t+1})w_{\hat{a}}(s_{t},s_{t+1}),\notag\\
w_{\hat{a}}(s_{t},s_{t+1}) &=\frac{P(s_{t+1}|s_t)}{\hat{a}(s_{t+1}|s_t)}.
\end{align}
Note that the importance weight $w_{\hat{a}}(s_{t},s_{t+1})$ requires access to the passive dynamics $P$.

\subsection{LMDPs With Transition-Dependent Rewards}

In the original formulation of LMDPs, reward is state-dependent. To develop a hierarchical framework based on LMDPs, we have to account for the fact that each task may accumulate different amounts of reward. Hence reward is transition-dependent, depending not only on the current state but also on the next state. In this section we extend LMDPs to transition-dependent reward, i.e.~the expected reward function $R:S\times S\rightarrow\mathbb{R}$ is defined over pairs of states. Then the reward $\mathcal{R}(s,a)$ of applying control $a$ in state $s$ is
\begin{align*}
\mathcal{R}(s,a)&=E_{s'}\left[R(s,s')\right]-\lambda\cdot\mathrm{KL}(a(\cdot|s)\Vert\, P(\cdot|s))\\
&=E_{s'}\left[R(s,s')-\lambda\cdot\log\frac{a(s'|s)}{P(s'|s)}\right].
\end{align*}
The Bellman equation becomes
\begin{align*}
\frac{1}{\lambda}V(s)&=\frac{1}{\lambda}\max_a\left\{\mathcal{R}(s,a)+E_{s'}\left[V(s')\right]\right\}\\
&=\max_aE_{s'}\left[\frac{1}{\lambda}(R(s,s')+V(s'))-\log\frac{a(s'|s)}{P(s'|s)}\right].
\end{align*}
Letting $Z(s)=e^{V(s)/\lambda}$ and $O(s,s')=e^{R(s,s')/\lambda}$ yields
\begin{align*}
\frac{1}{\lambda}V(s)&=-\min_aE_{s'}\left[\log\frac{a(s'|s)}{P(s'|s)O(s,s')Z(s')}\right].
\end{align*}
Normalizing as $\mathcal{G}[Z](s)=\sum_{s'}P(s'|s)O(s,s')Z(s')$ yields $V(s)/\lambda=\log\mathcal{G}[Z](s)$ and results in the policy
\[
a^*(s'|s)=\frac{P(s'|s)O(s,s')Z(s')}{\mathcal{G}[Z](s)}.
\]
Exponentiating the Bellman equation gives $Z(s)=\mathcal{G}[Z](s)$ which can be written in matrix form as
\begin{align}\label{eqn:trans-rew}
{\bf z}=\Gamma{\bf z},
\end{align}
where each entry of $\Gamma$ equals $\Gamma(s,s')=P(s'|s)O(s,s')$.

To solve Equation~\eqref{eqn:trans-rew} we can either apply the power iteration method or Z-learning. It is trivial to extend Z-learning to LMDPs with transition-dependent reward. Each transition is still a triplet $(s_t,r_t,s_{t+1})$, and the only difference is that the reward $r_t$ now depends on the next state $s_{t+1}$ as well as the current state $s_t$.
If we compare to the target value $Z(s)=G[Z](s)$, we see that
the update rule in Equation~\eqref{eqn:zlearning} causes $\hat{Z}$ to tend towards the optimal $Z$ value when using the uncontrolled distribution $P$ to sample transitions. The update for importance sampling in Equation~\eqref{eqn:zlearning-imp} can also be directly applied to LMDPs with transition-dependent reward.

\section{Hierarchical LMDPs}
\label{sec:hlmpds}


In this section we formalize a framework for hierarchical LMDPs based on MAXQ decomposition. We assume that there exists an underlying LMDP $L=\langle S,P,R\rangle$ and a set of tasks $\mathcal{L}=\{L_0,\ldots,L_n\}$, with $L_0$ being the root task. Each task $L_i=\langle T_i,A_i,\tilde{R}_i\rangle$, $0\leq i\leq n$, consists of a termination set $T_i\subset S$, a set of subtasks $A_i\subset\mathcal{L}$ and a pseudo-reward $\tilde{R}_i:T_i\rightarrow\mathbb{R}$. 
Each state $t\in T_i$ is absorbing and produces reward $\tilde{R}_i(t)$. For clarity of presentation, we first assume that each task $L_i$ is {\em deterministic}, i.e.~for each state $s\in S\setminus T_i$, $L_i$ terminates in a unique state $t_i(s)\in T_i$. We later show how to extend hierarchical LMDPs to non-deterministic tasks.


In MAXQ decomposition, since the actions of the original MDP $M=\langle S,A,P,R\rangle$ are included as primitive tasks, the action set $A_i$ of task $M_i$ contains a subset of actions in $A$. The analogy for hierarchical LMDPs is that each task $L_i$ contains a subset of the {\em allowed transitions} of the original LMDP $L=\langle S,P,R\rangle$, i.e.~transitions with non-zero probability according to $P$.
Intuitively, the optimal control $a_i$ of each task $L_i$ can be viewed as a stochastic policy that selects between deterministic tasks and deterministic next states.

We associate each task $L_i$ with an LMDP $\langle S,P_i,R_i\rangle$ with transition-dependent rewards.
 Task $L_i$ is {\em primitive} if and only if $A_i=\emptyset$. For each state $s\in S$, let $N_s\subseteq N(s,P)$ be the subset of next states of the original LMDP $L$ that are also present in $L_i$. Further let $A_s=\{L_j\in A_i\mid s\notin T_j\}$ be the set of subtasks of $L_i$ that are applicable in $s$, i.e.~for which $s$ is not a terminal state. Clearly, $A_s=\emptyset$ if $L_i$ is primitive.




For a given state $s\in S$, the passive dynamics $P_i$ and immediate reward $R_i$ of the task LMDP $L_i$ are defined in terms of transitions due to $N_s$
\begin{align*}
P_i(s'|s)  &=\frac{P(s'|s)}{\sum_{s''\in N_s}P(s''|s)}\cdot\frac{|N_s|}{|N_s|+|A_s|}, & \forall & s'\in N_s,\\
R_i(s,s')  &=R(s),                                                                    & \forall & s'\in N_s,
\end{align*}
and transitions due to $A_s$
\begin{align}
P_i(t_j(s)|s) &=\frac{1}{|N_s|+|A_s|},                                                   & \forall & L_j\in A_s,\label{eq:pj}\\
R_i(s,t_j(s)) &=V_j(s),                                                                  & \forall & L_j\in A_s.\label{eq:rj}
\end{align}
Just as in MAXQ decomposition, the reward associated with applying a subtask $L_j\in A_i$ of $L_i$ in state $s$ equals the value function of $L_j$ in $s$, i.e.~$R_i(s,t_j(s))=V_j(s)$. The transition $(s,t_j(s))$ associated with subtask $L_j\in A_i$ has uniform probability, while transitions in $N_s$ have probabilities proportional to those of $P$ and produce the same reward as in the original LMDP $L$.



\begin{figure}[t]
  \begin{center}
  \includegraphics[scale=1]{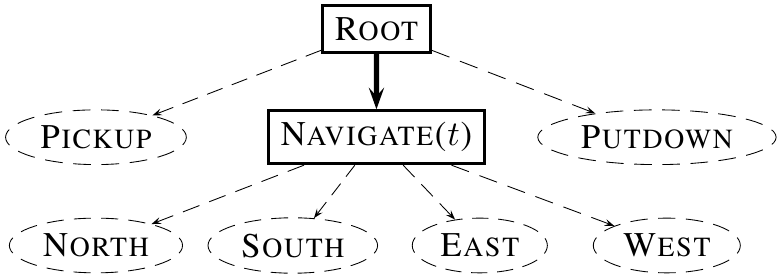}
  \end{center}

  \caption{The LMDP task graph of the Taxi domain.}
  \label{fig:taxi-lmdp}
\end{figure}

For each task $L_i$, the value function $V_i$ estimates the expected cumulative reward until $L_i$ terminates
and defines the immediate reward for higher-level tasks. 
We can write the Bellman optimality equation for $L_i$ as
\begin{align}\label{eqn:hier-lmdp}
\frac{V_i(s)}{\lambda}&=\max_{a_i}E_{s'}\left[\frac{1}{\lambda}(R_i(s,s')+V_i(s'))-\log\frac{a_i(s'|s)}{P_i(s'|s)}\right].
\end{align}

The task graph for hierarchical LMDPs is defined as for MAXQ decomposition, and has to be acyclic. Figure~\ref{fig:taxi-lmdp} shows the LMDP task graph of the Taxi domain.
Compared to Figure~\ref{fig:taxi}, primitive actions no longer appear as tasks, and new sink nodes, e.g.~\textsc{Navigate($t$)}, correspond to primitive tasks of the hierarchical LMDP. 

The above definitions implicitly consider the following assumptions that differ from the 
MAXQ formulation and are required for hierarchical LMDPs:
\begin{enumerate}
\item 
First, we assume that terminal states of subtasks are {\em mutually exclusive} and do not overlap with next states in $N_s$. The reason is that an LMDP is not allowed to have more than one transition between two states with {\em different} rewards: if this happens, the optimal policy is not identifiable, since one has to collapse both transitions into one and determine what the resulting immediate reward should be, an ill-defined problem.
\item 
Another difference is that each LMDP task $L_i$ needs the equivalent of a no-op action, i.e.~$P_i(s|s)>0$, so that the corresponding Markov chain is aperiodic, needed for convergence of the power-iteration method. 
\item 
Finally, unlike MAXQ decomposition, the value function $V_i$ in Equation~\eqref{eqn:hier-lmdp} includes KL terms due to differences between the control $a_i$ and uncontrolled dynamics $P_i$. The reward $R_i(s,t_j(s))=V_j(s)$ also includes subtask dependent KL terms. Consequently, the value function $V_0(s)$ of the root task $L_0$ includes KL terms from all other tasks. Although this introduces an approximation, we can control the relative importance of KL terms by adjusting the value of $\lambda$.
\end{enumerate}
\subsection{Task Compositionality for Non-Deterministic Tasks}
\label{sec:comp}

In this subsection, we extend the definition of hierarchical LMDPs to non-deterministic tasks.
As before, we associate with each task $L_i$ an LMDP $\langle S,P_i,R_i\rangle$ with the important difference that $L_i$ (and it subtasks) can have more than
one terminal state.
Primitive subtasks (or transitions $N_s$) are addressed as before, and we omit them for clarity.
We thus have to define passive dynamics and immediate rewards for non-primitive subtasks $L_j$ that can have more than one terminal state.

Denote $t_{j,k}(s)\in T_j$, $1\leq k\leq |T_j|$, as the $k$-th terminal state of subtask $L_j$.
We define the counterparts of Equations~\eqref{eq:pj} and~\eqref{eq:rj} for multiple terminal states as
\begin{align}
P_i(t_{j,k}(s)|s) & = \frac{\overline{P}_j(t_{j,k}(s)|s)}{|N_s|+|A_s|}, & \forall L_j&\in A_s,t_{j,k}(s)\in T_j,\label{eq:pjk}\\
R_i(s,t_{j,k}(s)) & = V_{j,k}(s),                         & \forall L_j&\in A_s, t_{j,k}(s)\in T_j.\label{eq:vjk}
\end{align}
where the transition probability $\overline{P}_j(t_{j,k}(s)|s)$ of subtask $L_j$ from state $s$ to terminal state $t_{j,k}(s)$ and the value function $V_{j,k}(s)$ 
can be expressed using compositionality of optimal control laws for LMDPs~\cite{TodorovNIPS2009}, as described below.
Note that $\overline{P}_j$ is different from the immediate transition probabilities $P_j$ for subtask $L_j$, and that the total transition probability of subtask $L_j$ is still $1/(|N_s| + |A_s|)$, but it is now distributed among the possible terminal states according to $\overline{P}_j(t_{j,k}(s)|s)$.

For each terminal state $t_{j,k}(s)\in T_j$ of task $L_j$, we define a separate task $L_{j,k}$ .
The new tasks are identical to $L_j$ and have $|T_j|$ terminal states that differ only in their pseudo-rewards $\tilde{R}_{j,k}$.
For task $L_{j,k}$, the pseudo-reward for $t_{j,k}(s)$ (goal) is zero and the remaining $|T_j|-1$ terminal states have the same (negative) pseudo-reward $C$,
e.g. $\tilde{R}_{j,k}(t_{j,k}(s))=0$ and $\tilde{R}_{j,k}(t_{j,l})=C$, $l\neq k$.

Consider the optimal policy $a^*_{j,k}(\cdot|s)$ and the optimal value $Z_{j,k}(s)$ of each of the individual tasks.
Using compositionality, the original task $L_j$ with multiple terminal states can be expressed as a weighted sum of the individual tasks $L_{j,k}$.
In particular, the composite optimal $Z_j$ and policy $a^*_{j}$ are~\cite{TodorovNIPS2009}
\begin{align*}
Z_j(s)           & = \frac{1}{|T_j|}\sum_{k=1}^{|T_j|}Z_{j,k}(s),\\
a^*_{j}(\cdot|s) & = \sum_{k=1}^{|T_j|} \frac{Z_{j,k}(s)}{Z_j(s)} a^*_{j,k}(\cdot|s),
\end{align*}
where the mixing weights for composing tasks are uniform and equals $1/|T_j|$, since each task $L_{j,k}$ assigns the same pseudo-reward $C$ to non-goal terminal states.

The value function in Equation \eqref{eq:vjk} is then given by 
\[V_{j,k}(s)= \lambda \log \frac{Z_{j,k}(s)}{Z_j(s)},\]
and the transition probability for Equation~\eqref{eq:pjk} is defined
recursively for all states $s$ as
\begin{align*}
\overline{P}_j(t_{j,k}(s)|t_{j,k}(s)) &= 1, &\\
\overline{P}_j(t_{j,k}(s)|t_{j,l}(s)) &= 0, & l\neq k,\\
\overline{P}_j(t_{j,k}(s)|s) &= \sum_{s'}a^*_{j}(s'|s)\overline{P}_j(t_{j,k}(s)|s'), & s \notin T_j.
\end{align*}
%
This defines the hierarchical LMDP framework for non-deterministic tasks.
Note that each individual task $L_{j,k}$ is still deterministic; its purpose is to {\em avoid} terminating in a state different from $t_{j,k}(s)$.

\subsection{Hierarchical Learning Algorithms}

The aim of a hierarchical LMDP is to learn an estimated hierarchical control policy $\hat{a}=\langle \hat{a}_0,\ldots,\hat{a}_n\rangle$, i.e.~an individual control policy $\hat{a}_i$ for each task $L_i$, $0\leq i\leq n$. Similar to MAXQ decomposition, there are two main alternatives for learning a hierarchical policy:
\begin{enumerate}
\item Learn each individual policy $\hat{a}_i$ separately in a bottom-up fashion.
\item Learn all policies simultaneously using a hierarchical execution in a top-down fashion.
\end{enumerate}
Implementing an algorithm of the first type is straightforward: since each individual task $L_i$ is an LMDP, we can using the power iteration method or Z-learning. Since all subtasks of $L_i$ are solved before $L_i$ itself, the rewards of $L_i$ are known and fixed when solving $L_i$.

To implement an algorithm of the second type, similar to MAXQ-Q learning, we start at the root task $L_0$ and sample a subtask $L_i$ to execute using the current estimate $\hat{a}_0$ of the policy for $L_0$. We then execute $L_i$ until termination, possibly applying subtasks of $L_i$ along the way. When $L_i$ terminates, we return the control to the root task $L_0$ and another subtask $L_j$ is sampled using $\hat{a}_0$. This continues until we reach an absorbing state of $L_0$. During this process, the value function estimates of each task are updated using Z-learning.

As in MAXQ-Q learning, if a task $L_i$ has pseudo-rewards different from $0$, we have to learn two estimates of the value function for $L_i$: one estimate $\hat{V}_i$ of the optimal value function $V_i$ 
that excludes the pseudo-reward $\tilde{R}_i$, and another estimate $\tilde{V}_i$ that includes the pseudo-reward $\tilde{R}_i$. The estimate $\tilde{V}_i$ defines the policy $\hat{a}_i$ of $L_i$, while $\hat{V}_i$ is passed as reward to parent tasks of $L_i$.

\subsection{Intra-Task Z-Learning}
\label{sec:intraZ}
In hierarchical MDPs, the aim is to learn a separate policy for each individual task. Since Q-learning is an off-policy algorithm, it is possible to use transitions recorded during one task to learn the policy of another task. Such intra-task learning is known to converge faster \cite{Sutton98intra-optionlearning}. In this section we describe an algorithm for intra-task Z-learning.

As described in Subsection~\ref{sec:lmdps}$\,\,\,\,$, we can use importance sampling to improve exploration.
Let $(s_t,r_t,s_{t+1})$ be a transition sampled using the estimated policy $\hat{a}_j$ of task $L_j$, and consider an update to the estimated value $\hat{Z}_i$ of another task $L_i$. 
Even though the sample distribution $\hat{a}_j$ is different from the estimated policy $\hat{a}_i$ of $L_i$, we consider the update in Equation~\eqref{eqn:zlearning-imp}
\begin{align}\label{eq:zil-hier}
\hat{Z}_i(s_t) \leftarrow (1-\alpha)\hat{Z}_i(s_t) + \alpha e^{r_t/\lambda}\hat{Z}_i(s_{t+1})w_{\hat{a}_i}(s_{t},s_{t+1}).
\end{align}
To see why the update rule is correct, simply substitute the expressions for $w_{\hat{a}_i}$ and $\hat{a}_i$:
\begin{align*}
\hat{Z}_i(s_t) \leftarrow~ & (1-\alpha)\hat{Z}_i(s_t) + \alpha e^{r_t/\lambda}\hat{Z}_i(s_{t+1})\frac{P(s_{t+1}|s_t)}{\hat{a}_i(s_{t+1}|s_t)}\\
 =~&(1-\alpha)\hat{Z}_i(s_t)\\& + \alpha e^{r_t/\lambda}\frac{P(s_{t+1}|s_t)\hat{Z}_i(s_{t+1})}{P(s_{t+1}|s_t)\hat{Z}_i(s_{t+1})}G[\hat{Z}_i](s_t)\\
 =~&(1-\alpha)\hat{Z}_i(s_t)\\& + \alpha e^{r_t/\lambda}G[\hat{Z}_i](s_t).
\end{align*}
In other words, instead of moving $\hat{Z}_i(s_t)$ in the direction of $e^{r_t/\lambda}\hat{Z}_i(s_{t+1})$, the update rule moves $\hat{Z}_i(s_t)$ in the direction of $e^{r_t/\lambda}G[\hat{Z}_i](s_t)$, which is precisely the desired value of $\hat{Z}_i(s_t)$. In particular, the importance weight can be shared by the different tasks in
intra-task learning. 

For LMDPs with transition costs, the same update rule can be used, but substituting the expressions for $w_{\hat{a}_i}$ and $\hat{a}_i$ leads to a slightly different result:
\begin{align*}
\hat{Z}_i(s_t) \leftarrow~& (1-\alpha)\hat{Z}_i(s_t) + \alpha e^{r_t/\lambda}\hat{Z}_i(s_{t+1})\frac{P(s_{t+1}|s_t)}{\hat{a}_i(s_{t+1}|s_t)}\\
 =~&(1-\alpha)\hat{Z}_i(s_t)\\& + \alpha\frac{P(s_{t+1}|s_t)e^{r_t/\lambda}\hat{Z}_i(s_{t+1})G[\hat{Z}_i](s_t)}{P(s_{t+1}|s_t)O(s_t,s_{t+1})\hat{Z}_i(s_{t+1})}\\
 =~&(1-\alpha)\hat{Z}_i(s_t)\\& + \alpha E[G[\hat{Z}_i](s_t)].
\end{align*}
Recall that $O(s_t,s_{t+1})=e^{R(s_t,s_{t+1})/\lambda}$. The expectation $E[G[\hat{Z}_i](s_t)]$ results from the fact that the observed reward $r_t$ and expected reward $R(s_t,s_{t+1})$ may be different, but are equal in expectation. For LMDPs with transition costs, $G[\hat{Z}_i](s_t)$ is the desired value of $\hat{Z}_i(s_t)$.

\subsection{State Abstraction}
\label{sec:abs}

In hierarchical LMDPs, we can apply the same forms of state abstraction as for MAXQ decomposition~\cite{conf/nips/Dietterich99}. The most common form of state abstraction is projection or {\em max node irrelevance}. This form of state abstraction assumes that the state is {\em factored}, i.e.~$S=S_1\times\cdots\times S_k$ where $S_1,\ldots,S_k$ are the domains of $k$ state variables. Max node irrelevance identifies state variables that are irrelevant for a given task, implying that the values of these state variables remain the same until completion of the task. Irrelevant state variables can be ignored while learning the value function.

\citeauthor{conf/nips/Dietterich99} (\citeyear{conf/nips/Dietterich99}) identified other conditions under which it is safe to perform state abstraction. One condition, leaf irrelevance, does not apply to hierarchical LMDPs since actions are no longer included as leaves of the task graph. Another condition, result distribution irrelevance, does apply to hierarchical LMDPs: when two or more states have the same transition probabilities with respect to a given task $L_i$, we only need to estimate a single value $\hat{V}_i$ (or $\tilde{V}_i$) for these states.

\begin{figure}[t]
\centering
\includegraphics[width=.78\columnwidth]{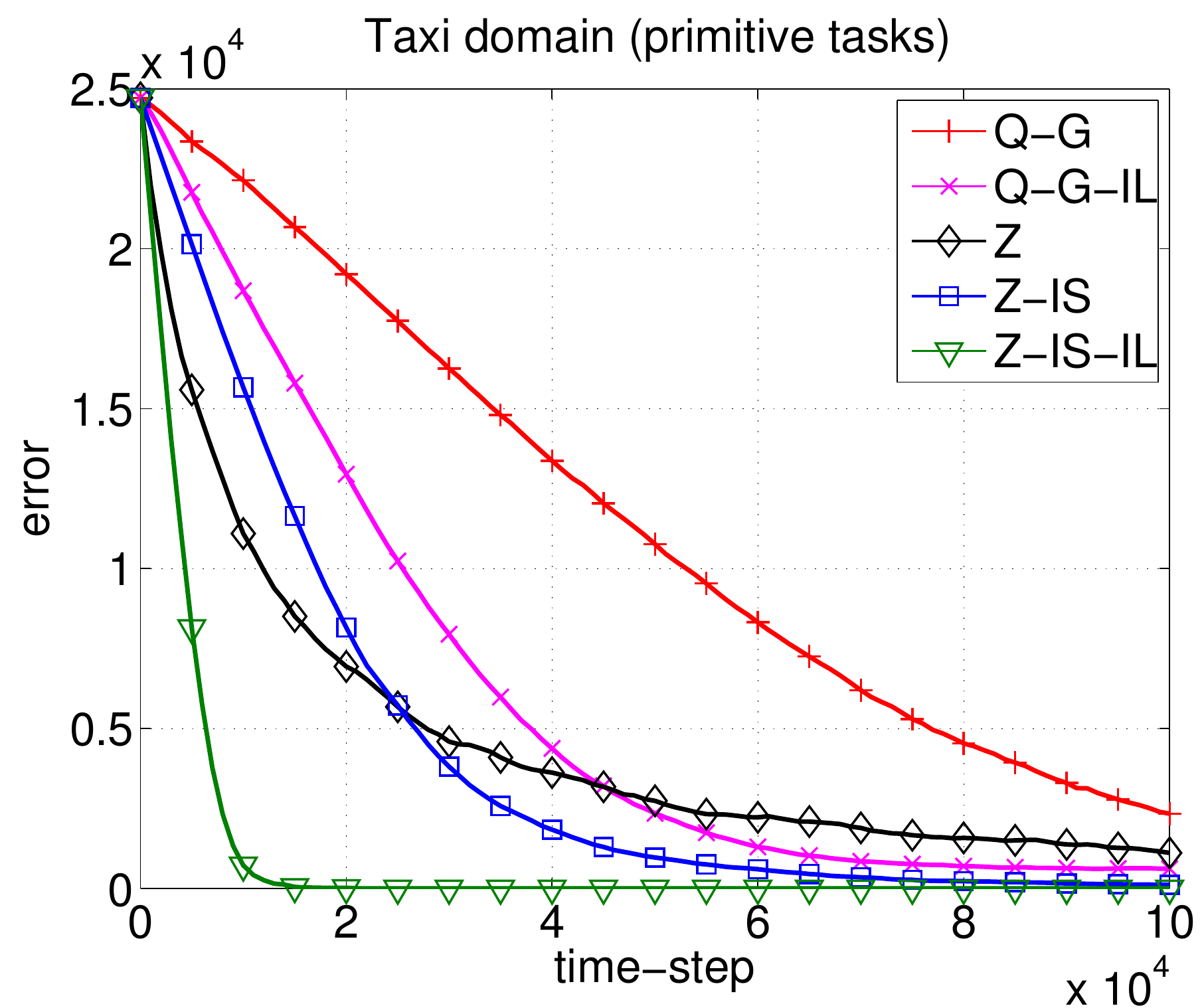}
\caption{Taxi problem: results the primitive tasks (NAVIGATE) for $\lambda=1$.}\label{fig:taxi_primitive}
\end{figure}

\section{Experiments}
\label{sec:exp}
We evaluate the proposed framework in two tasks commonly used in hierarchical MDPs: the taxi domain~\cite{dietterich2000hierarchical} and 
an autonomous vehicle guided task (AGV)~\cite{ghavamzadeh2007hierarchical}.
We compare the following methods:
\begin{description}
\item[Z:] Z-learning using naive sampling (i.e.~random walk) without correction term, as in Equation~\eqref{eqn:zlearning}.
\item[Z-IS:] Z-learning with importance sampling but without intra-task learning, as in Equation~\eqref{eqn:zlearning-imp},
\item[Z-IS-IL:] Z-learning with importance sampling and intra-task learning, as in Equation~\eqref{eq:zil-hier}.
\item[Q-G:] $\epsilon$-greedy Q-learning without intra-task learning.
\item[Q-G-IL:] $\epsilon$-greedy Q-learning with intra-task learning.
\end{description}

The Z-learning variants are evaluated using task LMDPs as described in Section~\ref{sec:hlmpds}~~
To compare with the Q-learning variants, for each task LMDP we construct a traditional MDP following the methodology of \citeauthor{TodorovPNAS2009}~\shortcite{TodorovPNAS2009}.
The resulting traditional MDP is guaranteed to have the same optimal value function as the original LMDP.
Following \citeauthor{TodorovNIPS2007}~\shortcite{TodorovNIPS2007}, we use dynamic learning rates, which decay as $\alpha(\tau)=c/(c + \tau)$, where $c$ is optimized separately for each algorithm and $\tau$ is the current trial.
The parameter $\epsilon$ for Q-learning is also optimized for best performance.

To compare performance, we calculate, for each iteration, the $\ell_1$-norm of the differences between the learned and the optimal value function,
which can be computed exactly in the tasks we consider here.
More details about the experiments are described in the supplementary material.
\subsection{The Taxi Domain}
\begin{figure}[t]
\centering
\includegraphics[width=.76\columnwidth]{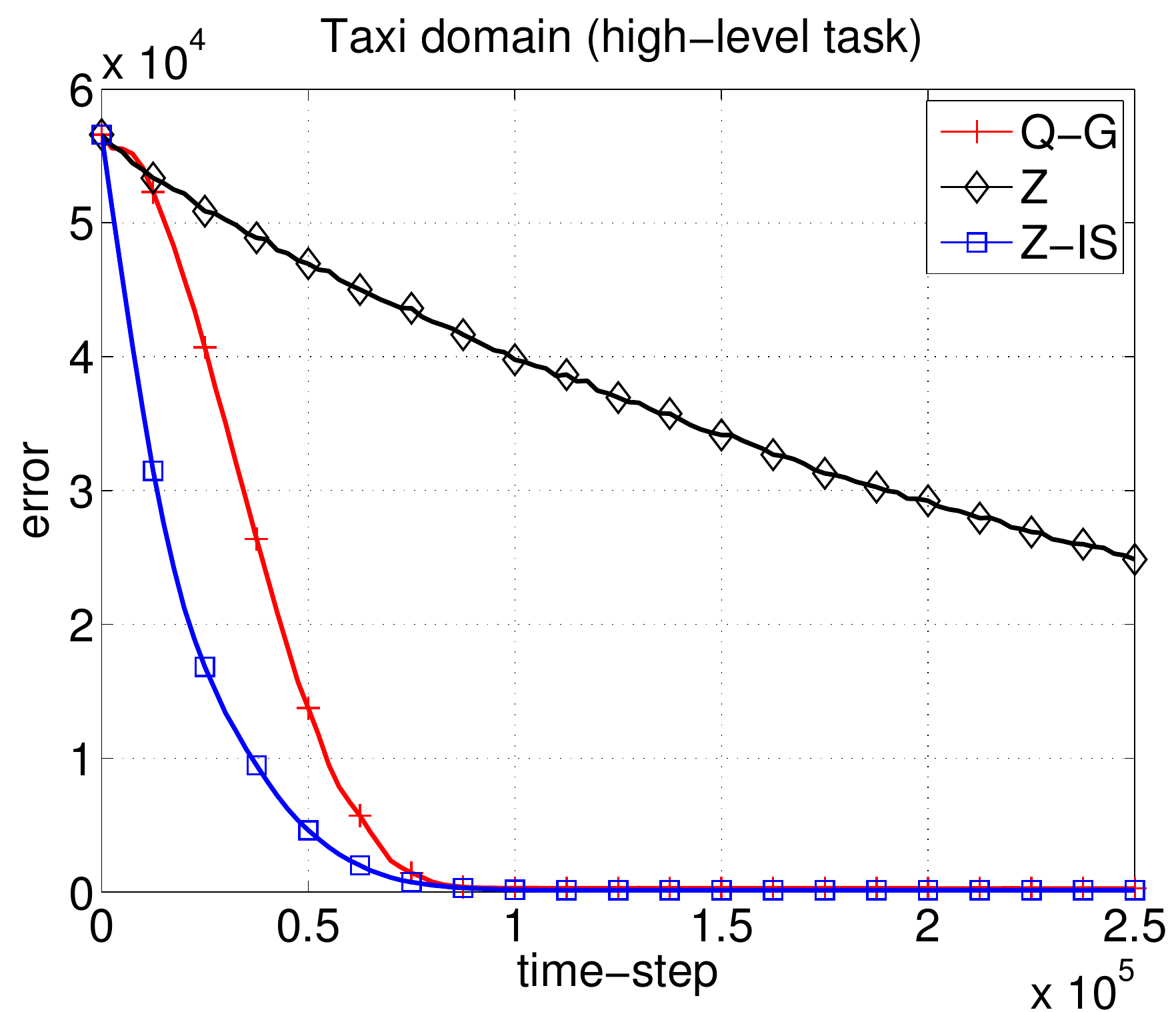}
\caption{Taxi problem: results in the abstract task (ROOT) for $\lambda=1$.}\label{fig:taxi_abstract}
\end{figure}
The taxi domain is defined on a grid, with four distinguished locations.
There is a passenger at one of the four locations, and that passenger wishes to be transported to one of the other three locations.
There is also a taxi that must navigate to the passenger's location, pick up the passenger, navigate to the destination location, and put down the passenger there.
We use a variant of the taxi domain \cite{dietterich2000hierarchical} with much larger state space, a $15\times 15$ grid.

We decompose the taxi domain as shown in Figure~\ref{fig:taxi-lmdp}.
Just like \citeauthor{dietterich2000hierarchical} (\citeyear{dietterich2000hierarchical}), we apply state abstraction in the form of projection to the navigate tasks, ignoring the passenger's location and destination.
This results in state spaces of sizes $625$ and $3{,}125$ for the navigation and full task, respectively.

The primitive tasks (NAVIGATE) contain all state transitions associated with navigation actions: NORTH, SOUTH, EAST, WEST and IDLE (i.e.~the no-op action).
There are four of these primitive tasks, one for each location (corner) in the grid.
The corresponding LMDPs are very similar to the grid example of \citeauthor{TodorovNIPS2007} (\citeyear{TodorovNIPS2007}):
the passive dynamics is a random walk and the state-cost term is zero for the terminal states (the corresponding corner)
and $1$ elsewhere.

Figure~\ref{fig:taxi_primitive} shows the performance of different learning methods in the primitive task of this domain.
The best results are obtained with Z-learning with importance sampling and intra-task learning (Z-IS-IL).
The Z-learning variants outperform the Q-learning variants mainly because Z-learning, unlike Q-learning, does not need a maximization operator
or state-action values~\cite{TodorovNIPS2007}.
Remarkably, Z-IS (without intra-task learning) still performs better than Q-learning with intra-task learning.
Naive Z-learning (Z) performs better than greedy Q-learning (Q-G) because in this particular task, random exploration 
is still useful to learn a location at one corner of the grid. 
\begin{figure}[t]
\centering
\includegraphics[width=.78\columnwidth]{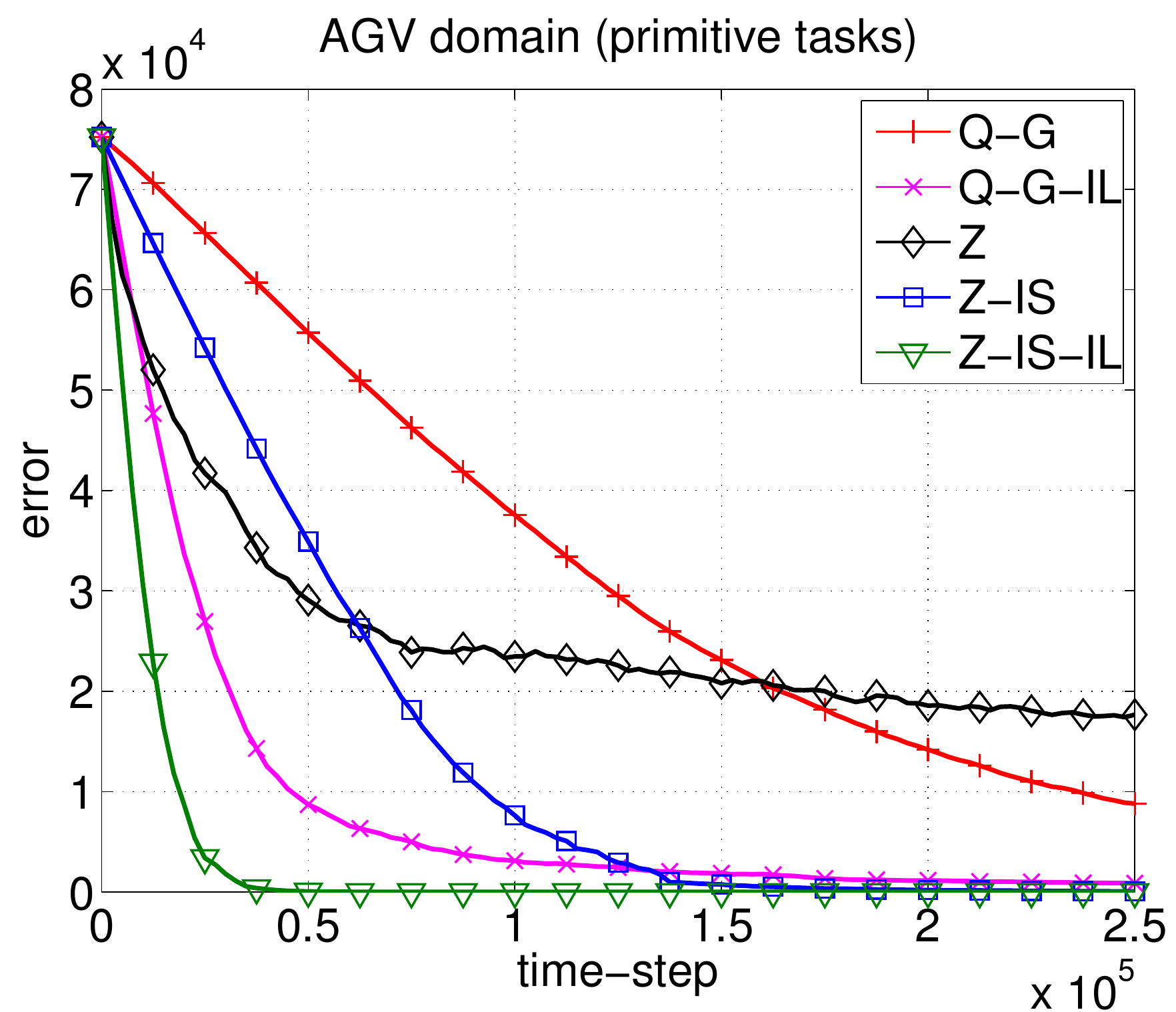}
\caption{AGV problem: results in the primitive tasks (NAVIGATE) for $\lambda=1$.}\label{fig:agvtasks}
\end{figure}

The full task is composed of the four possible navigation subtasks plus the transitions resulting from applying the original
PICKUP and PUTDOWN actions and the IDLE transition.
Figure~\ref{fig:taxi_abstract} shows the results in the full task.
Since there is only one such task, intra-task learning does not apply.
In this case, random exploration converges very slowly, as the curve of Z indicates.
Also, the advantage of Z-learning with importance sampling over $\epsilon$-greedy Q-learning is less pronounced
than in the primitive tasks.

From these results, we can conclude that the proposed extensions of Z-learning
outperform the state-of-the-art learning methods for this domain.


\subsection{The Autonomous Guided Vehicle (AGV) Domain}
The second domain we consider is a variant of the AGV domain \cite{ghavamzadeh2007hierarchical}.
In this problem, an AGV has to transport parts between machines in a warehouse.
Different parts arrive to the warehouse at uncertain times, and these parts can be loaded from
the warehouse and delivered to the specific machines that can process and assemble them.
Once a machine terminates, the AVG can pick up the assembly and bring it to the unload location
of the warehouse.

The state space of the full problem has nine components: three components for the position of the AGV ($x,y$ and angle),
one for the type of the part that the AGV is carrying and five to represent the different parts that are available to pick up from the warehouse of from the assembly locations.
To convert the overall problem into a first-exit task, we do not allow new parts to arrive at the warehouse, and the task is to assemble all parts and deliver them to the unload station.
For more details, see the supplementary material.

An important feature of this problem is that the AVG can only navigate using transitions
corresponding to primitive actions FORWARD, TURN-LEFT, TURN-RIGHT and STAY. 
Unlike the taxi domain, this significantly constrains the trajectories required to navigate from one location to another in the warehouse.
Similar to the taxi domain, we define six primitive NAVIGATE tasks for navigating to the six dropoff and pickup locations in the warehouse. As before, we apply state abstraction in the form of projection to these tasks, ignoring the location of all parts and assemblies.

Figure~\ref{fig:agvtasks} shows the result of different learning methods on the NAVIGATE tasks.
They are similar to those in the taxi domain, although Q-learning with intra-task learning performs comparatively better, while naive Z-learning performs comparatively worse.
The latter result can be explained by the need of guided exploration for navigating in this domain.

Since the total number of states is large, we also apply state abstraction in the form of result distribution irrelevance to the overall task. Since NAVIGATE tasks always terminate in a predictable state, it is not necessary to maintain a value function for other than dropoff and pickup locations.

We have also implemented an online algorithm similar to MAXQ-Q learning.
Instead of using the value function of subtasks to estimate transition rewards,
we execute each subtask until termination, recording the reward along the way.
The reward accumulated during the subtask is then used as the observed immediate reward for the abstract task.
For performance we measure the throughput, i.e.~the number of assemblies delivered to the unload station per time step.
\begin{figure}[t]
\centering
\includegraphics[width=.81\columnwidth]{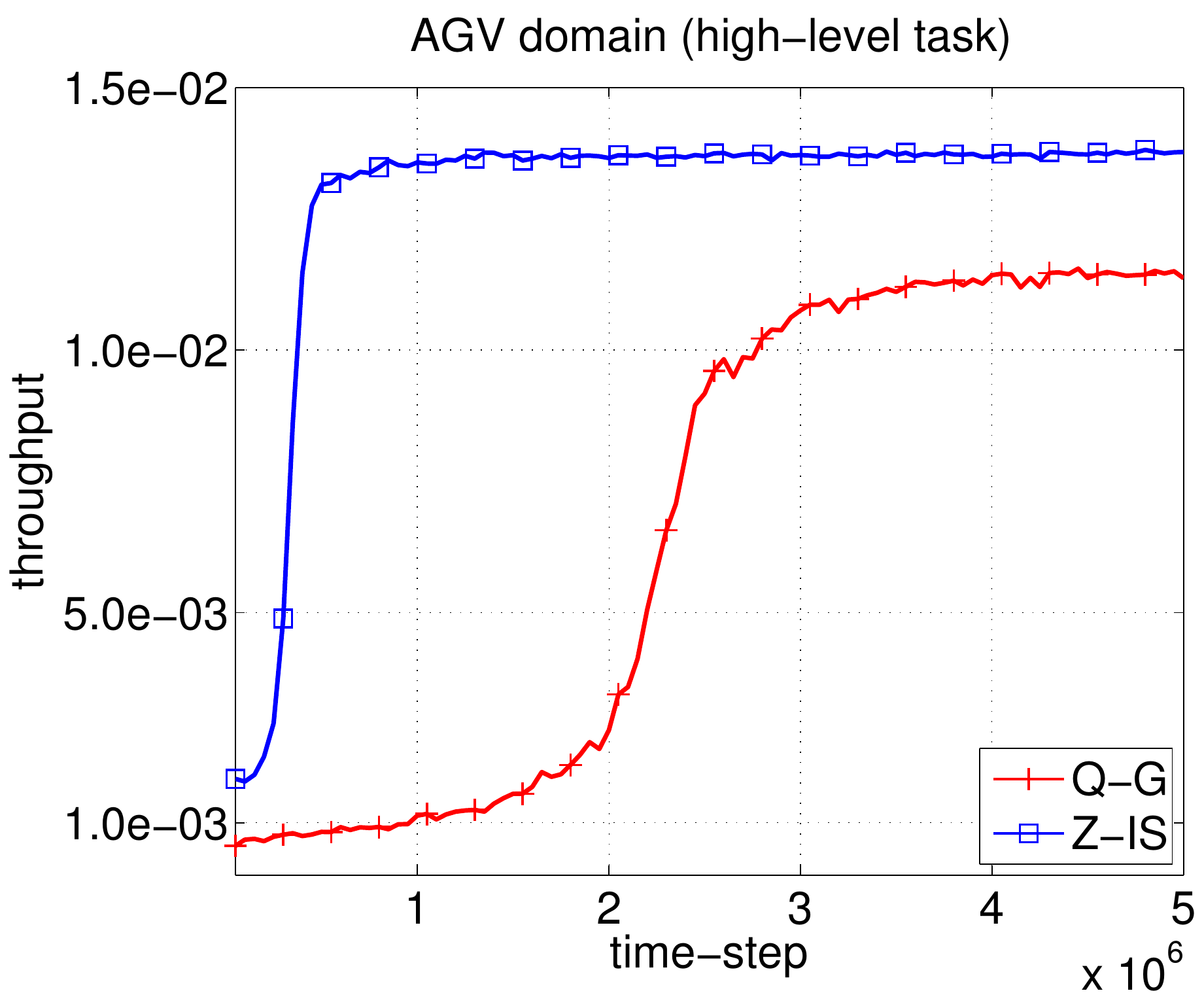}
\caption{AGV problem: comparison between Q-Learning and Z-learning in terms of throughtput.}\label{fig:agv}
\end{figure}
Figure~\ref{fig:agvtasks} shows the relative performance of Z-learning with importance sampling and Q-learning for this MAXQ-Q variant. We omit naive Z-learning, since the throughput of a random walk is constant over time. The number of time steps includes all primitive transitions, including those of the NAVIGATE subtasks. As the figure shows, Z-learning converges more quickly to a suboptimal policy compared to Q-learning, illustrating the benefits of hierarchical LMDPs.

\section{Conclusions}
\label{sec:end}
We have presented a framework for hierarchical reinforcement learning that combines the MAXQ decomposition
and formulates each task as a linearly-solvable MDP.
The framework has been illustrated in two domains, in which
the hierarchical, intra-task Z-learning algorithm outperforms the state-of-the-art methods for hierarchical MDPs.
 
%


\clearpage

\title{Hierarchical Linearly-Solvable Markov Decision Problems\\ (Supplementary Material)}
\author{Anders Jonsson \and Vicen\c{c} G\'omez\\
Department of Information and Communication Technologies\\
Universitat Pompeu Fabra\\
Roc Boronat 138, 08018 Barcelona, Spain\\
\{anders.jonsson,vicen.gomez\}@upf.edu
}
\maketitle

\appendix{}
\section{Experimental Setup}
To compare with the Q-learning variants, for each task LMDP we construct a traditional MDP following the methodology of \citeauthor{TodorovPNAS2009}~\shortcite{TodorovPNAS2009}.
For each state $s$, we define a symbolic action with transition probability distribution matching the optimal
$a^*(\cdot|s)$, which is computed using power iteration method in the original LMDP.
We also define as many other symbolic actions as number of possible states following $s$.
Their transition probabilities are obtained from $a^*(\cdot|s)$ by circular shifting.
The reward of the symbolic actions is $R(s) +\lambda \text{KL}(a^*(\cdot|s)||P(\cdot|s))$.
The resulting traditional MDP is guaranteed to have the same optimal value function that the original LMDP.

Following \citeauthor{TodorovNIPS2007}~\shortcite{TodorovNIPS2007}, we use dynamic learning rates, which decay as $\alpha(\tau)=c/(c + \tau)$, where the constant $c$ is optimized separately for each algorithm and $\tau$ is the current trial.
The parameter $\epsilon$ for Q-learning is also optimized for best performance.

To compare performance, we calculate, for each iteration, the $\ell_1$-norm of the differences between the learned and the optimal value function,

\subsection{The Taxi Domain} 
\begin{figure}[t]
\centering
\includegraphics[width=.8\columnwidth]{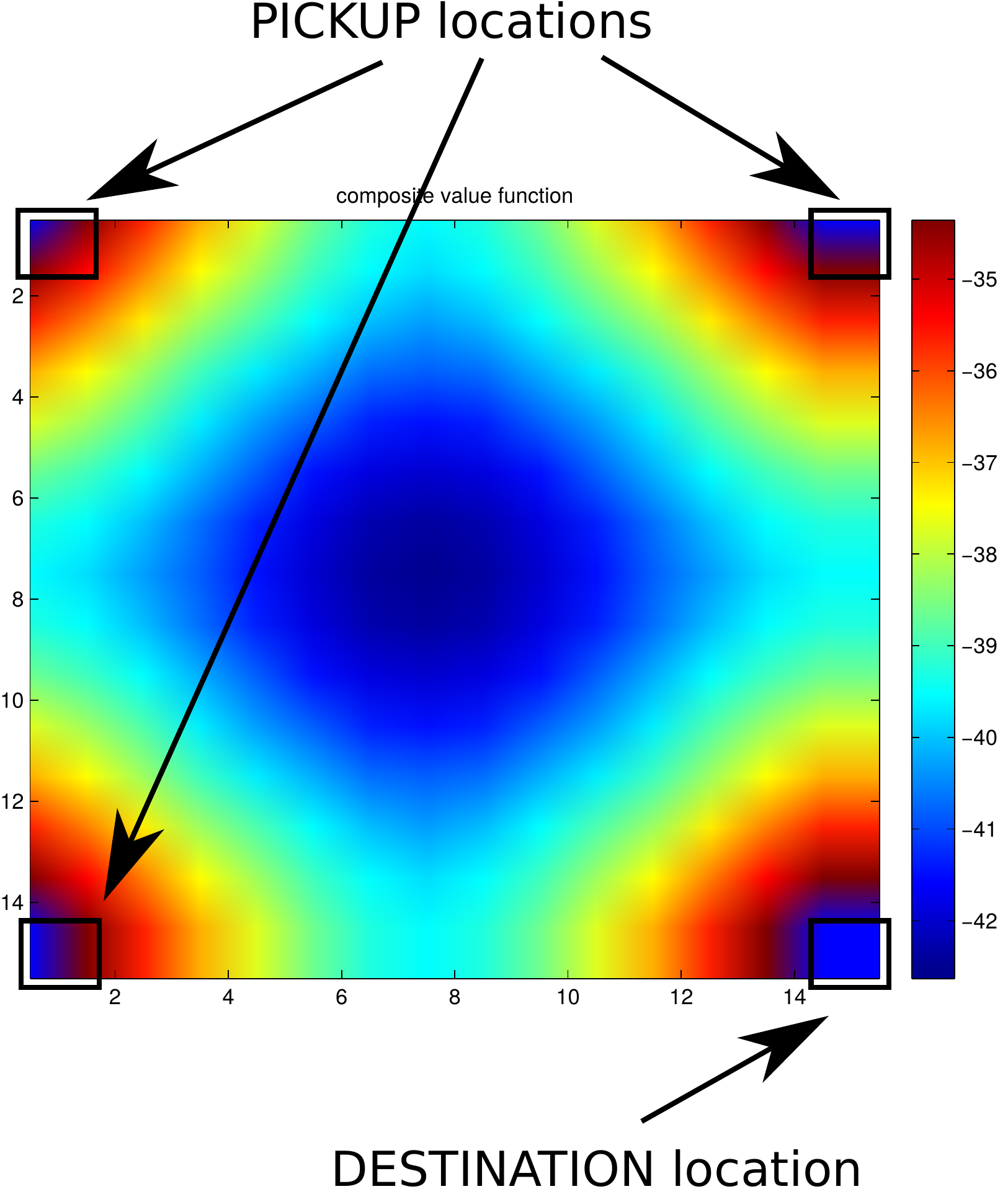}
\caption{Taxi domain used for the experiments.
In color, the value function of the composite task NAVIGATE.
}\label{fig:taxi_domain}
\end{figure}

The taxi domain is defined on a grid, with four distinguished locations.
There is a passenger at one of the four locations, and that passenger wishes to be transported to one of the other three locations.
There is also a taxi that must navigate to the passenger's location, pick up the passenger, navigate to the destination location, and put down the passenger there.
We use a variant of the taxi domain \cite{dietterich2000hierarchical} with much larger state space, a $15\times 15$ grid, as shown in Figure~\ref{fig:taxi_domain}.

The state space is composed of $(x,y,c)$, where $x$ and $y$ are the horizontal and vertical coordinates of the taxi location and $c=0,\hdots,4$
is the location of the passenger $0,\dots,3$ for the different pickup locations and $4$ when the passenger is in the taxi.
Just like \citeauthor{dietterich2000hierarchical} (\citeyear{dietterich2000hierarchical}), we apply state abstraction in the form of projection to the navigate tasks, ignoring the passenger's location and destination.
This results in state spaces of sizes $625$ and $3{,}125$ for the navigation and full task, respectively.
  
The primitive tasks (NAVIGATE) are composed of all state transitions corresponding to navigation actions, namely, NORTH, SOUTH, EAST, WEST and IDLE (i.e.~the no-op action).
There are four of these primitive tasks, one for each location (corner) in the grid.
The corresponding LMDPs are very similar to the grid example of \citeauthor{} (\citeyear{TodorovNIPS2007}):
the passive dynamics is a random walk and the state-cost term is zero for the terminal states (the corresponding corner)
and $1$ elsewhere.

In terms of scalability, the bottleneck of the algorithm is the numerical precision
required for computing the exact optimal value function.
This precision strongly depends on the absolute difference between the maximum and the minimum values 
in the matrices $\Pi$ (or $\Gamma$). In order to obtain good estimates,
one needs to set $\lambda$ sufficiently small, as mentioned in \citeauthor{TodorovNIPS2007} (\citeyear{TodorovNIPS2007}),
which increases the required numerical precision.
Although we obtained correct policies for larger grids, 
For $50\times 50$ grids we started to have numerical problems, since the threshold required to check convergence 
of the power iteration method is $\approx 10^{-300}$.

\subsection{The Autonomous Guided Vehicle (AGV) Domain}
\begin{figure}[t]
\centering
\includegraphics[width=.9\columnwidth]{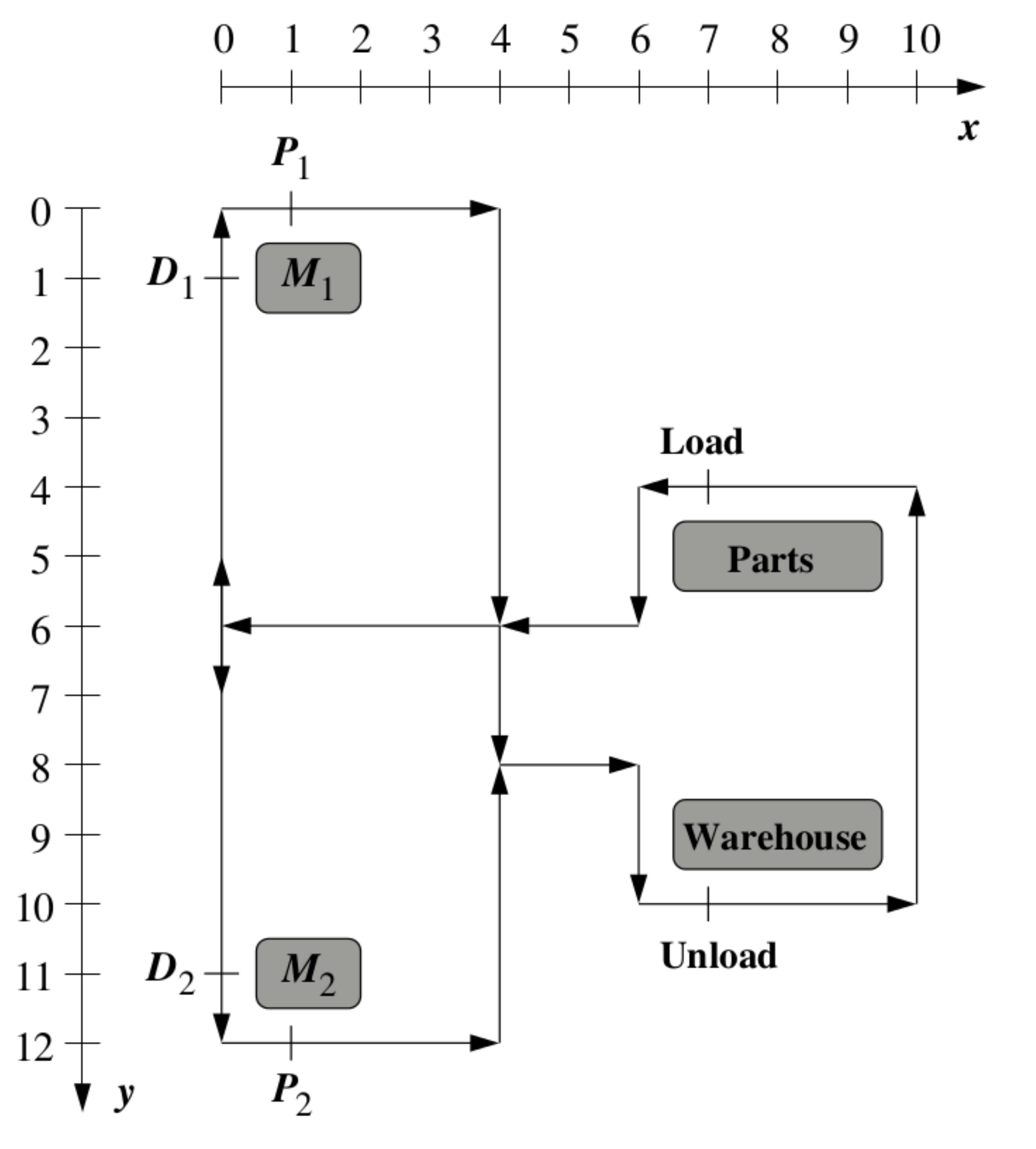}
\caption{AGV domain used for the experiments.}\label{fig:agv_domain}
\end{figure}
The second domain we consider is a variant of the AGV domain \cite{ghavamzadeh2007hierarchical}.
In this problem, an AGV has to transport parts between machines in a warehouse.
Different parts arrive to the warehouse at uncertain times, and these parts can be loaded from
the warehouse and delivered to the specific machines that can process and assemble them.
Once a machine terminates, the AVG can pick up the assembly and bring it to the unload location
of the warehouse.

We simplified the original problem by reducing the number of machines from $4$ to $2$ and setting the processing time of machines to
$0$ to make the task fully observable (see Figure \ref{fig:agv_domain}).

The state space of the full problem has the following components:
\begin{itemize}
\item $x$ coordinate of the AGV position
\item $y$ coordinate of the AGV position
\item $c$ orientation of the AGV (up,right,down,left)
\item Num. parts at the input buffer of Machine 1 $(0,1,2)$
\item Num. parts at the output buffer of Machine 1 $(0,1,2)$
\item Num. parts at the input buffer of Machine 2 $(0,1,2)$
\item Num. parts at the output buffer of Machine 2 $(0,1,2)$
\item Part of type 1 available at the warehouse $(0,1)$
\item Part of type 2 available at the warehouse $(0,1)$
\end{itemize}
To convert the overall problem into a first-exit task, we do not allow new parts to arrive at the warehouse, and the task is to assemble all parts and deliver them to the unload station.
The resulting task has approximately $75,000$ states.

The bottleneck of the algorithm is again a numerical precision error for calculating the optimal value function using the power iteration method.

\bibliography{bibtexdb.bib}
\bibliographystyle{aaai}

\end{document}